\def\year{2022}\relax
\documentclass[letterpaper]{article} 
\usepackage{aaai22}  
\usepackage{times}  
\usepackage{helvet}  
\usepackage{courier}  
\usepackage[hyphens]{url}  
\usepackage{graphicx} 
\usepackage{multirow}
\usepackage{subfigure}

\usepackage{amssymb}
\usepackage{cite}   
\usepackage{natbib}  
\usepackage{caption} 
\DeclareCaptionStyle{ruled}{labelfont=normalfont,labelsep=colon,strut=off} 
\usepackage{algorithm}
\usepackage{algorithmic}

\usepackage{color}

\newcommand{\ie}{\textit{i}.\textit{e}.,}

\usepackage{newfloat}
\usepackage{listings}
\floatstyle{ruled}
\newfloat{listing}{tb}{lst}{}
\floatname{listing}{Listing}

\title{Multi-Centroid Representation Network  for  Domain Adaptive Person Re-ID}
\author{
Yuhang Wu$^1$\thanks{This work was done when Yuhang Wu was an intern at Megvii Technology.}\thanks{equal contribution.}, Tengteng Huang$^2$\footnotemark[\value{footnote}], Haotian Yao$^2$, Chi Zhang$^2$,Yuanjie Shao$^1$,Chuchu Han$^1$, Changxin Gao$^1$, Nong Sang$^1$\thanks{Corresponding author.}, \\
	$^1$Key Laboratory of Ministry of Education for Image Processing and Intelligent Control, \\
	School of Artificial Intelligence and Automation, Huazhong University of Science and Technology\\
	$^2$Megvii Technology  \\
{\tt\small {\{wuyuhang,shaoyuanjie,hcc, cgao, nsang\}@hust.edu.cn }} \quad {\tt\small {tengtenghuang@foxmail.com}}\\
	{\tt\small {\{yaohaotian, zhangchi\}@megvii.com}} 
}
\usepackage{bibentry}

\begin{document}

 \maketitle

\begin{abstract}

Recently, many approaches tackle the Unsupervised Domain Adaptive person re-identification (UDA re-ID) problem through pseudo-label-based contrastive learning. During training, a uni-centroid representation is obtained by simply averaging all the instance features from a cluster with the same pseudo label. However, a cluster may contain images with different identities (label noises) due to the imperfect clustering results, which makes the uni-centroid representation inappropriate. In this paper, we present a novel Multi-Centroid Memory (MCM) to adaptively capture different identity information within the cluster. MCM can effectively alleviate the issue of label noises by selecting proper positive/negative centroids for the query image. Moreover, we further propose two strategies to improve the contrastive learning process. First, we present a Domain-Specific Contrastive Learning (DSCL) mechanism to fully explore intra-domain information by comparing samples only from the same domain. Second, we propose Second-Order Nearest Interpolation (SONI) to obtain abundant and informative negative samples. We integrate MCM, DSCL, and SONI into a unified framework named Multi-Centroid Representation Network (MCRN). Extensive experiments demonstrate the superiority of MCRN over state-of-the-art approaches on multiple UDA re-ID tasks and fully unsupervised re-ID tasks.

\end{abstract}
\begin{figure}[!ht]
    \centering
    \includegraphics{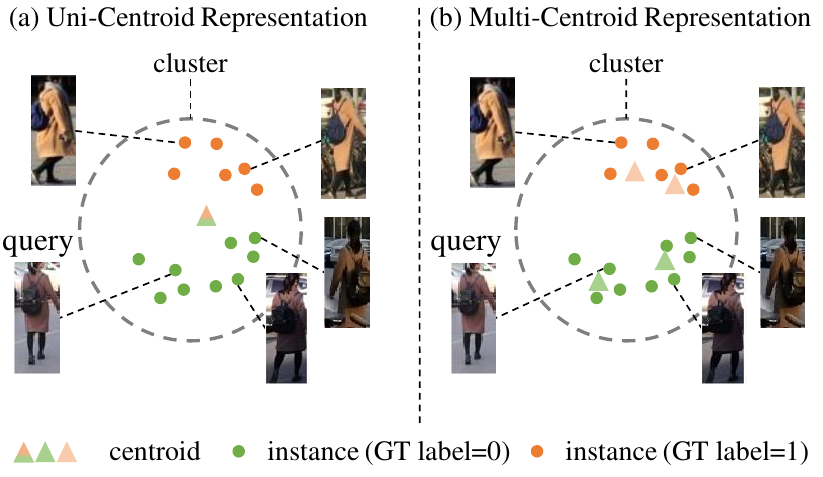}
    \caption{
    Comparison of traditional uni-centroid representation and our multi-centroid representation  when the cluster is mixed with different identities. (a) The uni-centroid representation incorporates multiple identity information which is inappropriate. (b) Our multi-centroid representation provides multiple discriminative centroids, making it possible to select a suitable centroid as the positive sample that captures the same identity information with the query.}
    
    \label{fig1}
\end{figure}
\section{Introduction}

Unsupervised Domain Adaptive person re-identification (UDA re-ID) is receiving increasing attention with the growing demand for practical video surveillance. The objective of UDA re-ID is to transfer knowledge learned from source domain with rich annotations to unlabeled target domain. Previous works usually tackle this problem by clustering~\cite{ge2020mutual,zhai2020ad,Ge2020spcl,zheng2021group,Zheng_Lan_Zeng_Zhang_Zha_2021}, which follow a two-step loop paradigm: (1) generating the pseudo labels of training samples from the target domain through clustering, (2) optimizing the model on the target domain with uni-centroid representation (\ie average feature or learnable weight of the cluster), under the supervision of the pseudo labels.

However, due to the imperfect results of the clustering algorithm, the pseudo labels always contain noises which are harmful to performance on the target domain.  For example, as shown in Figure~\ref{fig1}(a),  instances belonging to two identities are incorrectly merged into a cluster and assigned the same pseudo label. In this case, traditional uni-centroid representation inevitably incorporates information from different identities, which would mislead the feature learning  when the uni-centroid representation is used as the query's positive sample.

To alleviate the impact of label noises, we propose a Multi-Centroid Memory (MCM) to provide multiple centroids for each cluster. As Figure~\ref{fig1}(b) shows, each centroid  captures identity information within a local region of the cluster. This suggests that for each input query, we can select its reliable positive and negative samples from the centroids in the positive cluster and other negative clusters, respectively.
However, for a specific query, its positive centroids may contain some incorrect ones that capture different identity information with it. Such centroids will hinder the feature learning when used as the positive samples. To reduce the effect of these false-positive centroids, we propose a matching mechanism between the query and each positive centroid to select a centroid as the  positive sample. In general, the least similar positive centroid to the query is most likely be the false-positive centroid because of 
the unsatisfied inter-class separability and fixed clustering threshold, while the most similar one is not conducive to learning intra-class diversity. For striking a balance between the correctness and diversity, we select the moderate similar centroid  as the positive sample. Besides, the inferior clustering may also lead to some false-negative centroids, damaging the intra-class compactness. We select the mean negative centroid of each negative cluster as the negative sample, which is more reliable than using all negative centroids. 

In addition to considering the reliability of the positive and negative samples, we further considered their quality  for feature learning.
Some methods~\cite{Ge2020spcl,Zheng_Lan_Zeng_Zhang_Zha_2021,bai2021unsupervised} use valuable source domain data for training. Here, we follow these methods and extend our MCM to the source domain.
However, the cross-domain negative samples are quite easy for the query due to the huge domain gap between the source and target domains. These easy cross-domain negative samples contribute little to the optimization, and simply pushing them away from the query enlarges the domain gap. Given that, we propose Domain-Specific Contrastive Learning (DSCL) to fully mine intra-domain knowledge by only selecting the positive and negative samples from the query's domain for contrastive learning. Furthermore, inspired by the recent negative mining methods~\cite{kalantidis2020hard,zhong2021neighborhood} that use interpolation in the latent space to synthesize more negative samples, we propose  Second-Order Nearest Interpolation (SONI) to obtain 
additional hard negative samples for the query from the target domain. To ensure the synthetic negative samples are reliable and informative, SONI selects a set of nearest negative centroids and then uses each centroid as an anchor to interpolate with another nearest negative centroid that is nearest to it but has a different pseudo label. We integrate MCM, DSCL and SONI into a unified framework, Multi-Centroid Representation Network (MCRN), which provides each query with the positive and negative samples that are reliable and effective  for contrastive learning.

Our contributions can be summarized as follows:

\begin{itemize}

\item We propose a Multi-Centroid Memory (MCM) to alleviate the  label noise problem in previous UDA re-ID methods. By selecting reliable positive and negative centroids from MCM for each input query, the impact of label noises can be reduced.
\item  We further propose Domain-Specific Contrastive Learning (DSCL) and Second-Order Nearest Interpolation (SONI) to obtain  negative samples that are not only reliable but also effective for contrastive learning, which significantly improve the learning process.

\item Our integrated framework MCRN significantly outperforms state-of-the-art methods by a large margin on multiple UDA re-ID tasks. Besides,  extensive experiments on fully unsupervised re-ID tasks consistently demonstrate the superiority of our approach over previous methods.
\end{itemize}

\section{Related Work}
\subsection{Unsupervised domain adaptive (UDA) person re-ID}

The existing methods can be categorized into two branches, \ie, domain translation-based methods~\cite{wei2018person,deng2018image,zou2020joint} and clustering-based methods~\cite{ge2020mutual,zhai2020ad,Ge2020spcl,zheng2021group,Zheng_Lan_Zeng_Zhang_Zha_2021}. In this section, we mainly review clustering-based approaches since they are more related to our framework.

Clustering-based methods usually leverage the pseudo labels generated by clustering algorithms to optimize the network. However, it is quite challenging to assign correct pseudo labels to each unlabeled image due to the imperfect clustering results.  MMT~\cite{ge2020mutual} adopts a mutual mean-teaching framework to provide more robust soft labels.  NRMT~\cite{zhao2020unsupervised} performs collaborative clustering and mutual instance selection by maintaining two networks during training. UNRN ~\cite{Zheng_Lan_Zeng_Zhang_Zha_2021} introduces uncertainty estimation to explore the reliability of the pseudo label of each sample. SpCL~\cite{Ge2020spcl} propose the self-paced learning strategy to obtain more reliable clustering results. GLT~\cite{zheng2021group} uses a group-aware label transfer algorithm to online refine the pseudo labels. However, these works usually use average feature or learnable weight to represent a cluster, which is sensitive to label noises. Recently, a fully unsupervised re-ID method ClusterContrast~\cite{dai2021cluster} updates the cluster representation with the hardest positive instance feature in a batch, which is more robust than previous cluster representation.
Different from these methods, we introduce multiple centroids to adaptively detect and represent potential multiple sub-classes in a cluster.

\begin{figure*}[ht]
\centering
\includegraphics[]{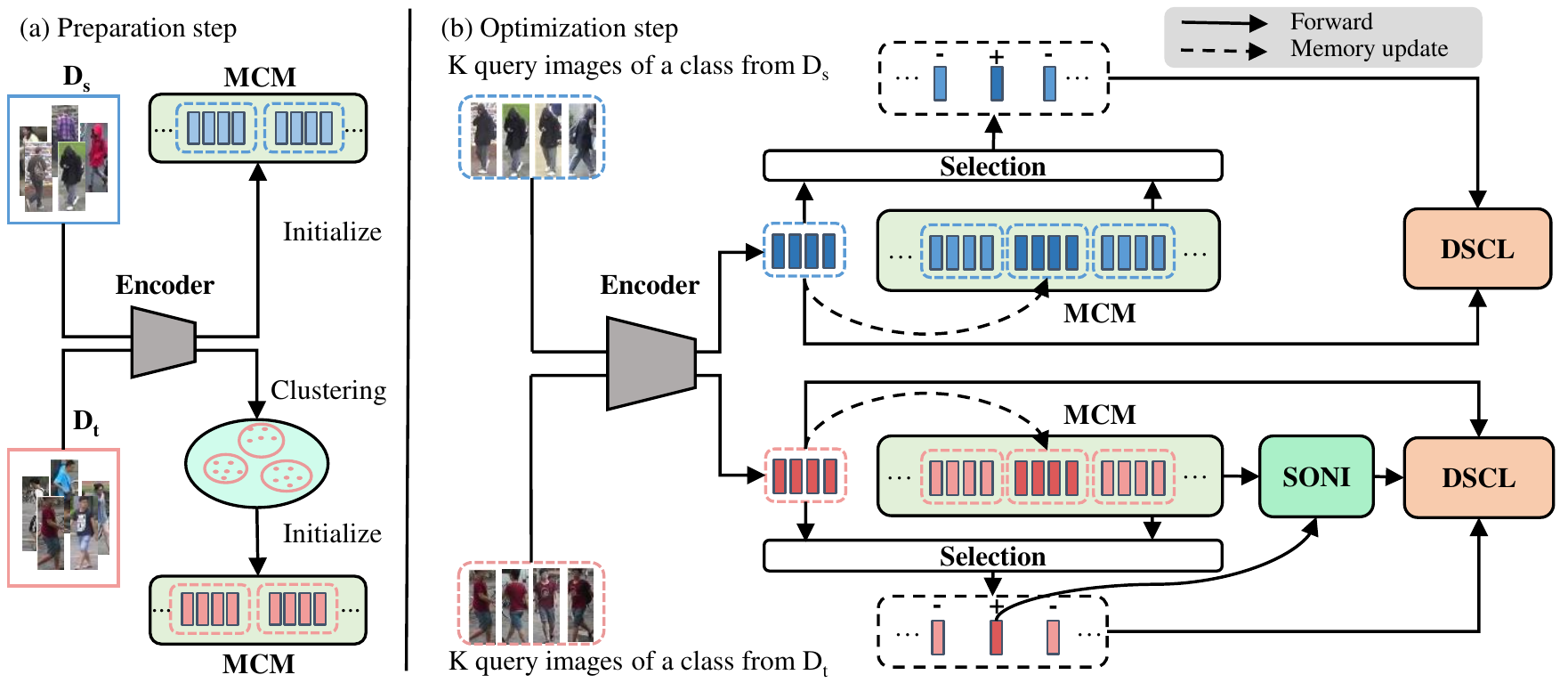} 
\caption{Illustration of the overall training pipeline of MCRN.  }
\label{framework}
\end{figure*}

\subsection{Contrastive Learning}
The contrastive loss~\cite{oord2018representation} is widely used in unsupervised visual representation learning task~\cite{chen2020simple,he2020momentum,tian2020contrastive} to learn discriminative feature representation by maximizing the similarity of augmented views generated from an identical instance.
Recently, SpCL~\cite{Ge2020spcl} successfully adapts contrastive loss to UDA re-ID task and propose a Unified Contrastive Loss~(UCL) to distinguish the query from negative samples from both the source and target domain. Different from UCL, we propose a novel Domain-Specific Contrastive Learning~(DSCL) mechanism which only selects informative negative samples from the same domain of query.

\subsection{Hard Negative Mining}
Mining hard negative samples plays an important role in boosting the performance of metric learning. The Embedding Expansion~\cite{ko2020embedding} employs uniform interpolation between two positive and negative points to generate many synthetic points and then select the hardest pair as negative.  NCD~\cite{zhong2021neighborhood} mixes the query in the novel classes with the samples in known classes to produce synthetic features, and then filters the hardest synthetic negatives by the cosine similarity with the query. MoCHi~\cite{kalantidis2020hard} synthesizes hard negatives by interpolating between the query and hard negative samples or any two randomly selected hard negative samples.
Unlike these methods, to ensure the reliability and quality of the  synthesized samples, we synthesize negative samples by interpolation between two hard negative centroids that are close to each other but have different pseudo labels.

\section{Proposed Method}

\subsection{Overview}

Given a source domain $D_{s}$ and a target domain $D_{t}$, the goal of UDA re-ID is to improve the model performance on $D_{t}$ by transferring knowledge from $D_{s}$ to $D_{t}$. ${D}_{s}=\left\{\left.\left({x}^{i}_{s},y^{i}_{s}\right)\right|_{i=1} ^{N_{s}}\right\}$ consists of $N_s$ labeled images, where $ \left({x}^{i}_{s},y^{i}_{s}\right)$ denotes the i-th training sample and its associated label. And ${D}_{t}=\left\{\left.\left({x}^{i}_{t}\right)\right|_{i=1} ^{N_{t}}\right\}$ composes of $N_{t}$ images without annotations.

In this paper, we propose a Multi-Centroid Representation Network~(MCRN), which consists of an encoder and a novel Multi-Centroid Memory (MCM). Moreover, we introduce a new Domain-Specific Contrastive Learning (DSCL) objective and a Second-Order Nearest Interpolation (SONI) mechanism to jointly improve the feature learning process. Figure~\ref{framework} illustrates the overall training pipeline of MCRN, which alternates between two steps: preparation step in Figure~\ref{framework}(a) and optimization step in Figure~\ref{framework}(b). In the preparation step, we group unlabeled images from $D_{t}$ into clusters using clustering algorithm~(\textit{e.g.}, DBSCAN~\cite{ester1996density}), based on the instance features extracted by the encoder. Then the Multi-Centroid Memory (MCM) is created and initialized, which is detailed later. In the optimization step, we carefully select positive/negative samples from MCM and generate more negative samples through SONI, followed by optimizing the encoder through DSCL between input queries and these samples. Note that MCM is dynamically updated during the optimization step.

\subsection{Multi-Centroid Memory}

\noindent\textbf{Memory initialization.}\ \ Each epoch begins with the preparation step. We first extract instance features for all the images from the source domain $D_s$ and the target domain $D_t$. Then we group the unlabeled target domain images into $n_t$ clusters through DBSCAN~\cite{ester1996density}.
In this process, we simply discard all the un-clustered instances. Then we build Multi-Centroid Memory (MCM) as a tensor in the shape of $M\times C$, where $M$ denotes the total number of centroids in MCM and $C$ denotes the dimension of feature channels. We set $M$ equal to $K\times \left ( n_{s}+n_{t} \right)$, where $K$ represents the number of centroids for each class. $n_{s}$ and $n_{t}$ denote the number of ground-truth classes from $D_s$ and pseudo classes~(\ie clusters) from $D_{t}$, respectively. We initialize all the $K$ centroids for a identical class as the mean feature of all the instance images from this class.

\noindent\textbf{Memory update.}\ \ In the optimization step, 
the centroids in MCM are continuously updated to detect and to represent potential multiple sub-classes with different identities, by continuously incorporating recent query features from the corresponding class. During training, MCM is update at the end of each iteration. Concretely, each mini-batch is composed of $P$ classes with $K$ queries per class, sampled by the widely used PK sampling approach. Note that $K$ is simply set identical to the number of centroids per class in MCM, which allows all of the centroids from the same class to be updated at the same pace. For the $K$ queries from the same class, we search for a permutation of corresponding $K$ centroids $\sigma \in \mathfrak{S}_{K}$ to establish best bipartite matching for query-centroid pairs with the highest overall similarity:

\begin{equation}
\hat{\sigma}= \mathop{\arg \max }_{{\sigma \in \mathfrak{S}_{K}}} \sum_{i}^{K} q^{i}\cdot c^{\sigma (i) }
\end{equation}
where $q^{i}$ denotes $i$-th sampled query of the class and $c^{\sigma (i)}$ indicates the $i$-th centroid in the permutation $\sigma$. We use Hungarian algorithm~\cite{kuhn1955hungarian} to efficiently compute the optimal permutation $\hat{\sigma}$. Based on the matched query-centroid pairs, we update the centroids in an exponential moving average manner as follows:
\begin{equation}\label{eq2}
    c^{\hat{\sigma} \left ( i \right ) } \leftarrow  mc^{\hat{\sigma} \left ( i \right ) }+\left (1-m  \right ) q^{i} 
\end{equation}
where $m$ is the momentum coefficient. We set $m$ to a relatively small value~(\textit{e.g.}, 0.2) so that the centroid can absorb more identity information from the query feature.

\subsection{Reliable centroids for contrastive learning}
For each query in the mini-batch, MCM provides $K$ positive candidates and  $(n_s+n_t-1)K$ negative candidates. How to obtain reliable and informative positive/negative samples plays an important role in the following contrastive learning.

\noindent\textbf{Reliable positive samples.}\ \ For a query, some of the $K$ positive candidates in MCM may capture different identity information~(\textit{i.e.}, false-positive centroid), due to incorrect clustering results. To obtain a reliable positive sample, we rank the $K$ positive candidates in ascending order according to their cosine similarity with the query. A natural choice is to select the candidate with the largest similarity as the positive sample. However, the most similar candidate usually incorporates the query feature in previous updates and is thus less informative for learning intra-class diversity. Besides, the least similar candidate is more likely to be an outlier. Therefore, we select the candidate ranked in the median (\ie $\left \lceil \frac{K}{2}  \right \rceil $), which we call the moderate positive centroid, as the positive sample.

\noindent\textbf{Reliable negative samples.}\ \ A naive choice is simply taking all the $(n_s+n_t-1)K$ negative candidates as negative samples. However, images with the same identity may be incorrectly split into multiple clusters due to unsatisfactory clustering results, resulting in false-negative candidates. Pushing the query and these false-negative candidates away would bias the feature learning. However, it is quite difficult to find and exclude possible false-negative candidates. To alleviate this problem, we represent each cluster as the mean feature of its $K$ centroids and take the mean feature~(named mean negative centroid)
as a negative sample. In this way, we can obtain $(n_t+n_s-1)$ in total negative samples from all the clusters except the one that the query falls in.

Notably, we use the same selection strategy for the query from $D_{s}$ to address the issue of possible annotation errors.

\subsection{Domain-Specific Contrastive Learning}

Previous work SpCL~\cite{Ge2020spcl} employs a Unified Contrastive Learning (UCL) to push samples from different classes away and pull those within the same class together. All the negative samples from the source and target domains are considered, no matter which domain the query comes from. UCL can be formulated as follows:

\begin{equation}\label{UCL}
    L_{U}=-\log \frac{\exp \left(\frac{1}{\tau} q \cdot c^{+}\right)}{\sum_{i=1}^{n_{s}} \exp \left(\frac{1}{\tau} q \cdot c_{s}^{i}\right)+\sum_{j=1}^{n_{t}} \exp \left(\frac{1}{\tau} q \cdot c^{j}_{t}\right)}
\end{equation}
where $q$ is the query in the mini-batch. $c^{i}_{s}$ and $c^{j}_{t}$ are the selected centroid of the $i$-th source-domain class and the $j$-th target-domain pseudo class, respectively. $c^{+}$ is the moderate positive centroid of the positive class and $\tau$ indicates the temperature coefficient.

However, due to the significant domain gap, it is quite easy for a model to distinguish the query from those negative centroids from a different domain. Such negative samples cannot provide effective information to learn discriminative representations.  Besides, simply pushing them away from the query enlarges the domain gap. Thus, we propose Domain-Specific Contrastive Learning (DSCL) which push query away from negative samples in the same domain:

\begin{equation}
    L_{D_{s}}=-\log \frac{\exp \left(\frac{1}{\tau} q_{s} \cdot c^{+}\right)}{\sum_{i=1}^{n_{s}} \exp \left(\frac{1}{\tau} q_{s} \cdot c^{i}_{s}\right)}
\end{equation}
\begin{equation}\label{lqt}
    L_{D_{t}}=-\log \frac{\exp \left(\frac{1}{\tau} q_{t} \cdot c^{+}\right)}{\sum_{i=1}^{n_{t}} \exp \left(\frac{1}{\tau} q_{t} \cdot c^{i}_{t}\right)}
\end{equation}
where $q_{s}$ and $q_{t}$ denote the query from source and target domains, respectively. Focusing on distinguishing the pairs from the same domain, DSCL can fully mine domain-specific semantic information and improve the generalization capability.

\subsection{Second-Order Nearest Interpolation}

We further propose a novel interpolation mechanism called Second-Order Nearest Interpolation (SONI) to synthesize abundant and informative negative samples. For each query from $D_t$, SONI interpolates between two hard negative centroids in MCM that are close to each other but belong to different pseudo classes. As indicated by its name, SONI involves two nearest neighbor searching processes. First, we collect the top $\gamma = \alpha n_t$ nearest negative centroids into a set $ H=\left \{ h^{j}|_{j=1}^{\gamma } \right \} $, where $\alpha$ is a hyper-parameter that controls the number of synthetic negative samples and $n_t$ is the number of pseudo classes. In this process, we use the moderate positive centroid to select the hard negative centroids since it is a reliable representation for the query. Then we search for the nearest negative neighbor $\tilde{h^{i}}\in H$ for each centroid $h^{i}\in H$. We interpolate between $h^{i}\in H$ and $h^{j}\in H$ to obtain a synthetic negative sample $s^{i}$.
\begin{equation}
    s^{i}= \beta h^{i}+\left ( 1-\beta  \right ) \tilde {h^{i}}
\end{equation}
where $\beta$ is randomly sampled from a uniform distribution in the range of $\left [ 0.2,0.5 \right ]$ in each iteration. 

We reformulate Equation~(\ref{lqt}) for DSCL as the follows to incorporate the negative samples generated by SONI:
\begin{equation}
    L_{D_{t}}^{*}=-\log \frac{\exp \left(\frac{1}{\tau} q_{t} \cdot c^{+}\right)  }{\sum_{i=1}^{n_{t}} \exp \left(\frac{1}{\tau} q_{t} \cdot c^{i}_{t}\right) +\sum_{j=1}^{\gamma} \exp \left(\frac{1}{\tau} q_{t} \cdot s^{j}\right)}
\end{equation}

where $s^{j}$ is the $j$-th synthetic negative sample for the query.

\subsection{Overall Loss}
Each mini-batch consists of $n$ encoded source-domain queries $Q_{s}=\left \{ q_s^{i}|_{i=1}^{n} \right \} $ and $n$ encoded target-domain queries $Q_{t}=\left \{ q_t^{i}|_{i=1}^{n} \right \} $. The overall optimization goal is as follows:

\begin{equation}
\centering
    L_{total}=\frac{1}{n}  \sum_{q_s\in Q_s}L_{D_{s}}+\frac{1}{n}  \sum_{q_t\in Q_t}L_{D_{t}}^*  
\end{equation}

\begin{table*}[!ht]
    \centering

\begin{tabular}{l|cc|cc|cc|cc}

\hline
\multicolumn{1}{c|}{\multirow{2}{*}{Methods}} & \multicolumn{2}{c|}{Duke$\to$Market} & \multicolumn{2}{c|}{Market$\to$Duke} & \multicolumn{2}{c|}{Duke$\to$MSMT} & \multicolumn{2}{c}{Market$\to$MSMT} \\
\multicolumn{1}{c|}{}                         & mAP               & R1               & mAP               & R1               & mAP              & R1              & mAP               & R1               \\ \hline
Baseline &73.1 	&87.8 	&62.1 	&77.6 	&19.1 	&44.5 	&18.2 	&43.0 

   \\
MCM                                      & 81.5              & 92.2             & 70.8              & 83.5             & 30.2             & 60.3            & 28.1              & 57.5             \\ 
MCM+DSCL                                & 82.4              & 92.8             & 71.4             & 84.0             & 31.1             & 61.5            & 29.7              & 59.5             \\
MCM+SONI                                  & 82.9              & 93.2             & 70.8              & 83.9               & 33.7             & 65.0              & 32.3              & 64.0             \\

MCRN                            & \textbf{83.8}     & \textbf{93.8}    & \textbf{71.5}     & \textbf{84.5}    & \textbf{35.7}    & \textbf{67.5}   & \textbf{32.8}     & \textbf{64.4}    \\ \hline
\end{tabular}
\caption{Ablation studies of our proposed components. Baseline: uni-centroid baseline based on SpCL~\cite{Ge2020spcl}.}\label{components}
\end{table*} 
\section{Experiments}

\subsection{Datasets and Evaluation Metrics}
We evaluate our method on three person re-ID datasets, including Market-1501~\cite{zheng2015scalable}, DukeMTMC-reID ~\cite{ristani2016performance} and MSMT17~\cite{wei2018person}. Rank-1/5/10 (R1/R5/R10) of Cumulative Matching Characteristic (CMC) and mean average precision (mAP) are adopted  for evaluation.

\subsection{Training details of MCRN}

\noindent\textbf{Baseline.}\ \ We use SpCL~\cite{Ge2020spcl} as our uni-centroid baseline and follow its most settings. ResNet-50~\cite{he2016deep} pretrained on ImageNet is used as the backbone for our encoder. We adopt domain-specific BNs~\cite{chang2019dsbn} for narrowing the domain gap. DBSCAN~\cite{ester1996density} clustering followed by a self-paced strategy~\cite{Ge2020spcl} is adopt for generating pseudo labels. For a fair comparison of uni-centroid and multi-centroid settings, we make two modifications to the original SpCL. First, we reinitialize the memory bank at the beginning of every epoch, while SpCL only initializes once at the first epoch. Second, we simply discard un-clustered instances while SpCL keeps them.

\noindent\textbf{Training details.}\ \ Each mini-batch consists of 64 source domain images and 64 target domain images, with 4 images per ground-truth/pseudo class~(\textit{i.e.}, $K$ is set to 4). All training images are resized to 256$\times$128 and various data augmentations are applied, including random cropping, random flipping and random erasing~\cite{zhong2020random}. Adam optimizer is utilized to optimize the encoder with a weight decay of 0.0005. The initial learning rate is set to 0.00035 and is decayed by 1/10 every 20 epochs in the total 50 epochs. The momentum coefficient $m$ in Equation~\ref{eq2} is set to 0.2, and the temperature coefficient $\tau$ in the contrastive losses is set to 0.05. $\alpha$ in SONI is set to 0.03. We implement our approach using the Pytorch~\cite{paszke2019pytorch} framework and use four NVIDIA RTX-2080TI GPUs for training. 

\subsection{Ablation Studies}

\noindent\textbf{Superiority of multi-centorid representation.}\ \ In Table~\ref{components}, we compare the performance of uni-centroid~(\textbf{Baseline}) and multi-centroid~(\textbf{MCM}) representation method. Notably, both of them adopt UCL~(Equation \ref{UCL}) as the learning objective and the only difference between them is the representation of each class. The result shows that our MCM significantly surpasses the baseline by considerable margins. As shown in Table~\ref{components}, MCM outperforms baseline by 8.4$\%$/4.4$\%$, 8.7$\%$/5.9$\%$, 11.1$\%$/15.8$\%$ and 9.9$\%$/14.5$\%$ in terms of mAP/R1 on four UDA tasks, clearly demonstrating the superiority of multi-centroid representation over traditional uni-centroid representation.

\begin{table}[t]

\begin{tabular}{c|c|cc|cc}
\hline
\multirow{2}{*}{Positive} & \multirow{2}{*}{Negative} & \multicolumn{2}{c|}{Duke$\to$Market} & \multicolumn{2}{c}{Market$\to$Duke} \\
                          &                           & mAP               & R1               & mAP               & R1               \\ \hline

Most                      & Mean                      & 56.3              & 76.0             & 44.6                & 58.7               \\
Least                     & Mean                       & 79.9              & 91.2             & 69.4              & 82.8           \\ \hline
Moderate                  & All                      &81.0 &91.4 &70.0 &83.3      \\ \hline
Moderate                  & Mean                       & \textbf{81.5}     & \textbf{92.2}    & \textbf{70.8}     & \textbf{83.5}      \\ \hline
\end{tabular}

\caption{Comparison of different strategies for selecting positive/negative samples. 
}\label{cpadh}
\end{table}

\begin{table}[t]
\centering
\begin{tabular}{l|cc|cc}
\hline
\multicolumn{1}{c|}{\multirow{2}{*}{Methods}} & \multicolumn{2}{c|}{Duke$\to$Market}              & \multicolumn{2}{c}{Market$\to$Duke}              \\
\multicolumn{1}{c|}{}                         & \multicolumn{1}{c}{mAP} & \multicolumn{1}{c|}{R1} & \multicolumn{1}{c}{mAP} & \multicolumn{1}{c}{R1} \\ \hline
QNNI                                           & 76.5                    & 89.4                    & 66.1                    & 80.9                    \\
RNNI                                          & 83.4                    & 93.5                    & 70.4                    & 83.6                    \\
SONI                                          & \textbf{83.8}           & \textbf{93.8}           & \textbf{71.5}           & \textbf{84.5}           \\ \hline
\end{tabular}
\caption{Comparison with different strategies for synthesizing negative samples in MCRN.}\label{SONI}
\end{table}

\begin{table}[t]
\centering
\begin{tabular}{c|c|c|c|c|c}
\hline
\multirow{2}{*}{Value of K} & \multicolumn{5}{c}{Duke$\to$Market} \\  \cline{2-6} 
                            & 2     & 3     & 4     & 5     & 6    \\ \hline \hline
mAP                         & 75.7  & 83.2  & \textbf{83.8}  & 83.5  & 83.5 \\
R1                          & 89.3  & 93.0  & \textbf{93.8}  & 93.4  & 93.6 \\ \hline
\end{tabular}
\caption{Influence of the number of the centroids $K$ for each class.}\label{IoK}
\end{table}

\begin{table*}[!ht]
\centering
\begin{tabular}[\textwidth]{l|c|cccc|cccc}
\hline
\multicolumn{1}{c|}{\multirow{2}{*}{Methods}} & \multicolumn{1}{c|}{\multirow{2}{*}{Reference}} & \multicolumn{4}{c|}{DukeMTMC$\to$Market1501}                                                         & \multicolumn{4}{c}{Market1501$\to$DukeMTMC}                                                        \\ \cline{3-10} 
\multicolumn{1}{c|}{}                         & \multicolumn{1}{c|}{}                           & \multicolumn{1}{c}{mAP} & \multicolumn{1}{c}{R1} & \multicolumn{1}{c}{R5} & \multicolumn{1}{c|}{R10} & \multicolumn{1}{c}{mAP} & \multicolumn{1}{c}{R1} & \multicolumn{1}{c}{R5} & \multicolumn{1}{c}{R10} \\ \hline
AD-Cluster~\cite{zhai2020ad}                                    & CVPR 20                                         & 68.3                    & 86.7                   & 94.4                   & 96.5                     & 54.1                    & 72.6                   & 82.5                   & 85.5                    \\
MMT~\cite{ge2020mutual}                                          & ICLR 20                                         & 71.2                    & 87.7                   & 94.9                   & 96.9                     & 65.1                    & 78.0                     & 88.8                   & 92.5                    \\
NRMT~\cite{zhao2020unsupervised}                                          & ECCV 20                                         & 71.7                    & 87.8                   & 94.6                   & 96.5                     & 62.2                    & 77.8                   & 86.9                   & 89.5                    \\
MEB-Net~\cite{zhai2020multiple}                                       & ECCV 20                                         & 76.0                      & 89.9                   & 96.0                     & 97.5                     & 66.1                    & 79.6                   & 88.3                   & 92.2                    \\
DG-Net++~\cite{zou2020joint}                                      & ECCV 20                                         & 61.7                    & 82.1                   & 90.2                   & 92.7                     & 63.8                    & 78.9                   & 87.8                   & 90.3                    \\
SPCL~\cite{Ge2020spcl}                                          & NIPS 20                                         & 76.7                    & 90.3                   & 96.2                   & 97.7                     & 68.8                    & \underline{82.9}                   & 90.1                   & 92.5                    \\
HGA~\cite{Zhang_Liu_Li_Guo_Duan_Long_Jin_2021}                                           & AAAI 21                                         & 70.3                    & 89.5                   & 93.6                   & 95.5                     & 67.1                    & 80.4                   & 88.7                   & 90.3                    \\
UNRN~\cite{Zheng_Lan_Zeng_Zhang_Zha_2021}                                          & AAAI 21                                         & 78.1                    & 91.9                   & 96.1                   & 97.8                     & 69.1                    & 82.0                     & \underline{90.7}                   & \underline{93.5}                    \\
GCL~\cite{chen2021joint}                                           & CVPR 21                                         & 75.4                    & 90.5                   & 96.2                   & 97.1                     & 67.6                    & 81.9                   & 88.9                   & 90.6                \\ 
GLT~\cite{zheng2021group}                                           & CVPR 21                                         & 79.5                    & 92.2                   & 96.5                   & 97.8                     & \underline{69.2}                    & 82.0                     & 90.2                   & 92.8                    \\
RDSBN+MDIF~\cite{bai2021unsupervised}                                    & CVPR 21                                         & \underline{81.5}                    & \underline{92.9}                   & \textbf{97.6}                   & \underline{98.4}                    & 66.6                    & 80.3                   & 89.1                   & 92.6                    \\

\hline
MCRN &This paper &\textbf{83.8}	&\textbf{93.8}	&\underline{97.5}	&\textbf{98.5}	&\textbf{71.5}	&\textbf{84.5}	&\textbf{91.7}	&\textbf{93.8}

\\ \hline \hline
\multicolumn{1}{c|}{\multirow{2}{*}{Methods}} & \multicolumn{1}{c|}{\multirow{2}{*}{Reference}} & \multicolumn{4}{c|}{DukeMTMC$\to$MSMT17}                                                         & \multicolumn{4}{c}{Market1501$\to$MSMT17}                                                        \\ \cline{3-10} 
\multicolumn{1}{c|}{}                         & \multicolumn{1}{c|}{}                           & \multicolumn{1}{c}{mAP} & \multicolumn{1}{c}{R1} & \multicolumn{1}{c}{R5} & \multicolumn{1}{c|}{R10} & \multicolumn{1}{c}{mAP} & \multicolumn{1}{c}{R1} & \multicolumn{1}{c}{R5} & \multicolumn{1}{c}{R10} \\ \hline
MMT~\cite{ge2020mutual}                                           & ICLR 20                                         & 23.3                    & 50.1                   & 63.9                   & 69.8                     & 22.9                    & 49.2                   & 63.1                   & 68.8                    \\
DG-Net++~\cite{zou2020joint}                                      & ECCV 20                                         & 22.1                    & 48.8                   & 60.9                   & 65.9                     & 22.1                    & 48.4                   & 60.9                   & 66.1                    \\
SpCL~\cite{Ge2020spcl}                                          & NIPS 20                                         & 26.5                    & 53.1                   & 65.8                   & 70.5                     & 26.8                    & 53.7                   & 65.0                     & 69.8                    \\

UNRN~\cite{Zheng_Lan_Zeng_Zhang_Zha_2021}                                          & AAAI 21                                         & 26.2                    & 54.9                   & 67.3                   & 70.6                     & 25.3                    & 52.4                   & 64.7                   & 69.7                    \\
HGA~\cite{Zhang_Liu_Li_Guo_Duan_Long_Jin_2021}                                           & AAAI 21                                         & 26.8                    & 58.6                   & 64.7                   & 69.2                     & 25.5                    & 55.1                   & 61.2                   & 65.5                    \\
GLT~\cite{zheng2021group}                                           & CVPR 21                                         & 27.7                    & 59.5                   & 70.1                   & 74.2                     & 26.5                    & 56.6                   & 67.5                   & 72.0                      \\

GCL~\cite{chen2021joint}                                           & CVPR 21                                         & 29.7                    & 54.4                   & 68.2                   & 74.2                     & 27.0                      & 51.1                   & 63.9                   & 69.9         \\
RDSBN+MDIF~\cite{bai2021unsupervised}                                    & CVPR 21                                         & \underline{33.6}                    & \underline{64.0}                     & \underline{75.6}                   & \underline{79.6}                     & \underline{30.9}                    & \underline{61.2}                   & \underline{73.1}                   & \underline{77.4}                    \\

\hline
MCRN &This paper &\textbf{35.7}	&\textbf{67.5}	&\textbf{77.9}	&\textbf{81.6}	&\textbf{32.8}	&\textbf{64.4}   	&\textbf{75.1}	&\textbf{79.2}
 \\
\hline

\end{tabular}
\caption{Comparison with state-of-the-art UDA person re-ID methods on common UDA benchmarks.}\label{SOTA}
\end{table*}
\begin{table*}[!ht]
\centering

\begin{tabular}{l|c|cc|cc|cc}
\hline

\multicolumn{1}{c|}{\multirow{2}{*}{Methods}} & \multirow{2}{*}{Reference} & \multicolumn{2}{c|}{Market1501} & \multicolumn{2}{c|}{DukeMTMC} & \multicolumn{2}{c}{MSMT17}   \\ \cline{3-8} 
\multicolumn{1}{c|}{}                        &                            & mAP            & R1             & mAP           & R1            & mAP           & R1            \\ \hline
SpCL~\cite{Ge2020spcl}                                          & NIPS 20                    & 73.1           & 88.1           & 65.3          & 81.2          & 19.1          & 42.3          \\
GCL~\cite{chen2021joint} &CVPR 21 & 66.8 &87.3 & 62.8 &82.9 &21.3 &45.7 \\
RLCC~\cite{zhang2021refining}                                          & CVPR 21                    & \underline{77.7}           & \underline{90.8}           & \underline{69.2}          & \underline{83.2}          & \underline{27.9}          & \underline{56.5}          \\ 

\hline
MCRN
& This paper                 & \textbf{80.8}  & \textbf{92.5}  & \textbf{69.9} & \textbf{83.5} & \textbf{31.2} & \textbf{63.6} \\
\hline
\end{tabular}
\caption{Comparison with state-of-the-art fully unsupervised person re-ID methods on person re-ID datasets. }\label{usl}
\end{table*}

\begin{figure}[t]
    \centering
    \includegraphics{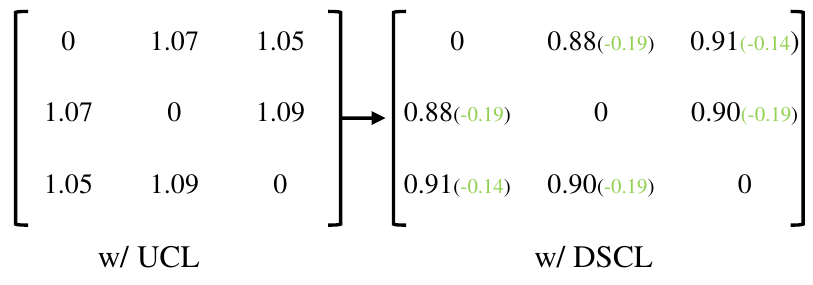}
    \caption{Pair-wise cosine distances among three domains:
Market, Duke and MSMT.}
    \label{DSACL}
\end{figure}

\noindent\textbf{Effectiveness of DSCL.}\ \ We further conduct experiments by replacing UCL with DSCL~(\textbf{MCM} v.s. \textbf{MCM+DSCL}). 
As Table~\ref{components} shows, DSCL brings the model with consistent performance  gain on all tasks. Specially, mAP/R1 is improved by 0.9$\%$/1.2$\%$ and 1.6$\%$/2.0$\%$ on Duke$\to$MSMT and Market$\to$MSMT tasks, respectively. Following \cite{bai2021unsupervised}, we compare the domain distance, which is measured by the cosine distance between the average feature of two domains. As is shown in Figure~\ref{DSACL}, DSCL reduces the distance between source and target domains, indicating the effectiveness of DSCL in reducing the domain gap.

\noindent\textbf{Effectiveness of SONI.}\ \ As is shown in Table~\ref{components}, SONI yields a general improvement on all the four UDA tasks. For example, mAP/R1  is increased by 3.5$\%$/4.7$\%$ and 4.2$\%$/6.5$\%$ on Duke$\to$MSMT and Market$\to$MSMT tasks, respectively. These results demonstrate the effectiveness of SONI in providing informative and beneficial negative samples. Besides, SONI is complementary to DSCL. When combine them together, our approach achieves superior results of mAP 83.8$\%$, 71.5$\%$, 35.7$\%$ and 32.8$\%$ on these tasks, respectively.

\subsection{Design Choices}
\noindent\textbf{Strategies for selecting positive samples.}\ \ Besides the moderate positive centroid, we further present two alternatives for selecting positive samples, \textit{i.e.}, selecting the most or the least similar centroid. We call these three strategies \textbf{Moderate}, \textbf{Most}, and \textbf{Least} for short. As is shown in Table~\ref{cpadh}, \textbf{Moderate} consistently outperforms \textbf{Most}/\textbf{Least} by a large margin, yielding an improvement of mAP 25.2$\%$/1.6$\%$ and 26.2$\%$/1.4$\%$ on Duke$\to$Market and Market$\to$Duke tasks, respectively. It might be surprising that \textbf{Most} leads to heavily degraded performance. We assume the reason is that the most similar centroid is likely to absorb the query feature in previous updates and thus is less informative for learning intra-class diversity.

\noindent\textbf{Strategies for selecting negative samples.}\ \ We further compare two strategies for selecting negative samples, including 1) mean negative centorid~(\textbf{Mean} for short) and 2) all negative centroids~(\textbf{All} for short).  For a query, \textbf{All} simply uses all centroids from a negative class, while \textbf{Mean} only takes the mean centroid. 
As is shown in Table~\ref{cpadh}, \textbf{Mean} leads to considerable and general improvements over \textbf{All}, which indicates that \textbf{Mean} can effectively filter bad samples and alleviate the issue of false-negative samples.

\noindent\textbf{MCM on the source data.} We ablate the effect of MCM on the source domain (MCM-S). On Duke$\to$Market, MCM-S outperforms Baseline~(in Table~\ref{components}) by +1.8\%/+1.1\% in terms of mAP/R1. We assume the reason is that MCM can increase the diversity of learned representation for each class. Therefore, we use MCM on both source and target domains.

\begin{figure}[!ht]
    \centering
    \includegraphics[width=0.45\textwidth]{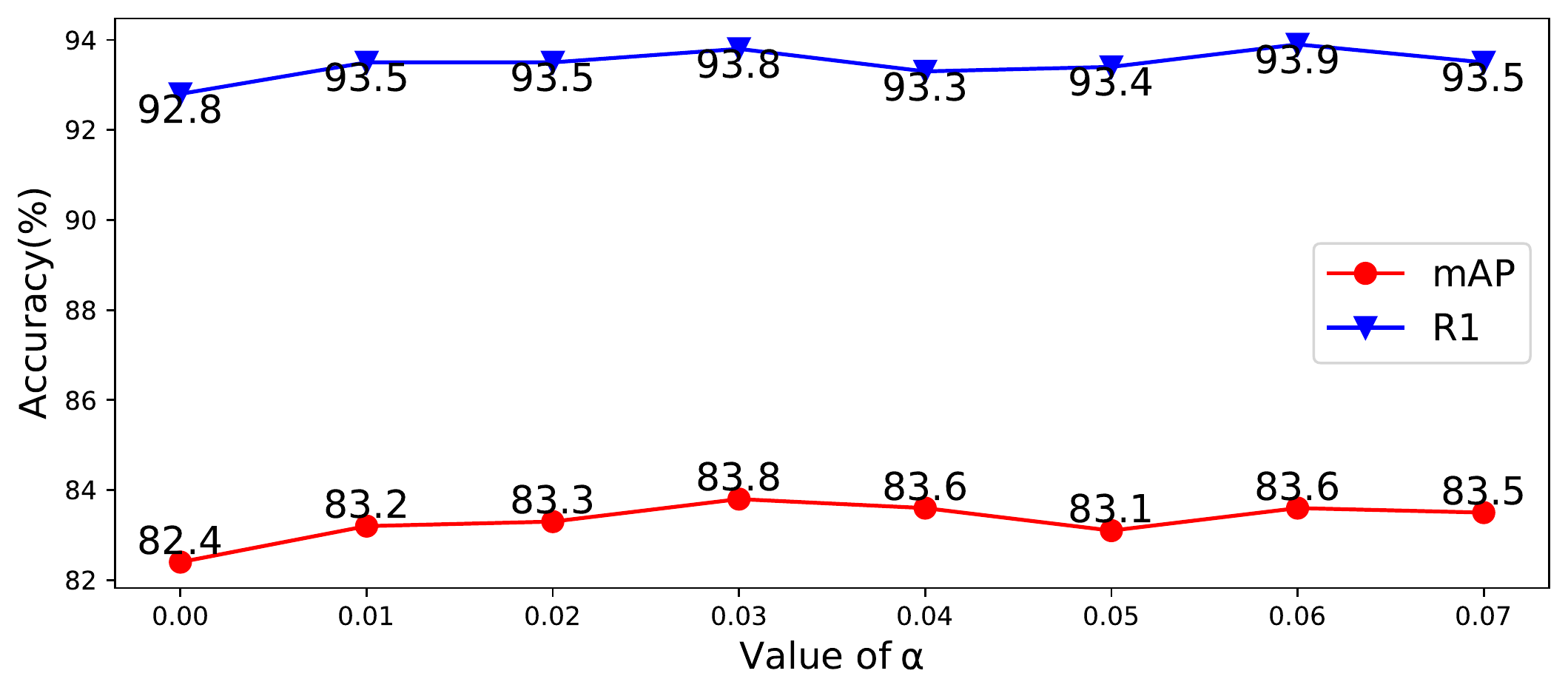}
    \caption{Performance of MCRN on Duke$\to$Market task with different value of $\alpha$.}
    \label{alpha}
\end{figure}

\noindent\textbf{Alternatives of interpolation approach.}\ \ Besides SONI, we evaluate another two interpolation methods proposed in MoCHi~\cite{kalantidis2020hard}. One is an interpolation between the query and its nearest negative sample~(QNNI for short). The other is an interpolation between two samples randomly selected from the top-$\gamma$ nearest negative samples~(RNNI for short). As is shown in Table~\ref{SONI}, SONI performs best among these three interpolation approaches. Since samples generated by QNNI incorporates the query feature, these samples are usually too hard to differentiate by the model and harmful for optimization. Instead of using two random negative samples adopted in RNNI, SONI interpolates between two negative samples which are semantically similar but have different pseudo labels, which is beneficial to obtain more informative samples.

\noindent\textbf{The number of centroids for each class.}\ \ We conduct experiments by varying $K$ from 2 to 6 with an interval of 1 to investigate how the number of centroids $K$ for each class influence the UDA performance.  As is shown in Table ~\ref{IoK}, the UDA performance is continuously boosted when $K$ is increased from 2 to 4. With larger $K$, the UDA performance reaches a plateau and no obvious gain is observed. Hence, we set $K=4$ as our default setting. 

\noindent\textbf{The number of synthetic negative samples.}\ \ We generate $\gamma=\alpha n_t$ synthetic hard negative samples through SONI. To explore the effect of the number of synthetic negative samples, we vary $\alpha$ from 0 to 0.07 with an interval of 0.01 and present the results in Figure~\ref{alpha}. As is shown, $\alpha$ in the range of $[0.01, 0.07]$ consistently outperforms $\alpha=0$, indicating the effectiveness of synthetic negative samples. Besides, with the 
increase of $\alpha$, the UDA performance first increases and then reaches a plateau, with the best performance achieved at 0.03. Hence, we set $\alpha=0.03$ as our default setting.


\subsection{Comparison with the State-of-the-arts}
\noindent\textbf{Performance under the UDA re-ID setting.}\ \ We compare our proposed MCRN with the state-of-the-art UDA re-ID methods on four domain adaptation tasks in Table~\ref{SOTA}. Our method  significantly outperforms the second best UDA re-ID methods by 2.3$\%$, 2.3$\%$, 2.1$\%$ and 1.9$\%$ in mAP on these tasks, respectively. With the same base configuration as SpCL, our method outperform SpCL by 7.1$\%$, 2.7$\%$,  9.2$\%$ and 6.0$\%$ in terms of mAP on these tasks, respectively. The comparison with MMT~\cite{ge2020mutual} and UNRN~\cite{Zheng_Lan_Zeng_Zhang_Zha_2021} are valuable, since they adopt a teacher-student framework which consists of two identical models while our method can outperform them with only a single model.

\noindent\textbf{Performance under the fully unsupervised re-ID setting.}\ \ Our proposed method can be easily generalized to fully unsupervised re-ID tasks. We compare our method with other state-of-the-art approaches for unsupervised re-ID in Table~\ref{usl}. As is shown, our method remarkably surpasses the state-of-the-art fully unsupervised person re-ID methods on Market, Duke and MSMT datasets, which validates the effectiveness of our method once again.  Specially, our MCRN outperforms the second best method RLCC~\cite{zhang2021refining} by 3.1$\%$/1.7$\%$ and 3.3$\%$/7.1$\%$ in mAP/R1 on Market and MSMT datasets, respectively.

\section{Conclusion}
In this work, we propose a unified framework, Multi-Centroid Representation Network (MCRN), to address the unsupervised domain adaptive person re-ID task. To alleviate the impact of label noises, we propose a Multi-Centroid Memory (MCM) to capture more identity information and select reliable positive/negative samples for each input query. In order to learn more discriminative feature representation, we propose Domain-Specific Contrastive Loss (DSCL) to fully explore intra-domain  information and Second-Order Nearest Interpolation (SONI) to enrich informative hard negative samples for the query from the target domain.  Extensive experiments have demonstrated the effectiveness of our framework.

\section{Acknowledgements}
This work was supported by the National Key R\&D Plan of the Ministry of Science and Technology (Project No. 2020AAA0104400), the Project of the National
Natural Science Foundation of China No. 61876210, the
Fundamental Research Funds for the Central Universities
No.2019kfyXKJC024, and the 111 Project on Computational
Intelligence and Intelligent Control under Grant
B18024.
\bibliography{aaai22}

\end{document}